\pdfoutput=1
\documentclass[11pt]{article}

\usepackage[review]{acl}
\usepackage{times}
\usepackage{latexsym}
\usepackage{graphicx}
\usepackage{color}
\usepackage{subcaption}
\usepackage[T1]{fontenc}
\usepackage[utf8]{inputenc}

\usepackage{microtype}

\usepackage{inconsolata}
\usepackage{amsmath}
\usepackage{amssymb}
\usepackage{booktabs}
\usepackage[symbol]{footmisc}

\title{SoftDedup: an Efficient Data Reweighting Method for Speeding Up Language Model Pre-training}

\author{\textbf{Nan He\thanks{~~Work done during internships at Tencent AI Lab.} \quad 
Weichen Xiong\footnotemark[1]\quad
Hanwen Liu\footnotemark[1]\quad
Yi Liao\thanks{\llap{}\:\:\:Corresponding author.}\quad 
Lei Ding}  \\
\textbf{Kai Zhang\thanks{\llap{}\:\:\:Kai Zhang proposed the initial prototype of the method.}\quad
Guohua Tang\quad
Xiao Han\quad  Wei Yang
}\\
        Tencent AI Lab\\
	henan991201@gmail.com, leoeliao@tencent.com
}
\begin{document}
\maketitle
\begin{abstract}
The effectiveness of large language models (LLMs) is often hindered by duplicated data in their extensive pre-training datasets. Current approaches primarily focus on detecting and removing duplicates, which risks the loss of valuable information and neglects the varying degrees of duplication. To address this, we propose a soft deduplication method that maintains dataset integrity while selectively reducing the sampling weight of data with high commonness. Central to our approach is the concept of "data commonness", a metric we introduce to quantify the degree of duplication by measuring the occurrence probabilities of samples using an n-gram model. Empirical analysis shows that this method significantly improves training efficiency, achieving comparable perplexity scores with at least a 26\% reduction in required training steps. Additionally, it enhances average few-shot downstream accuracy by 1.77\% when trained for an equivalent duration. Importantly, this approach consistently improves performance, even on rigorously deduplicated datasets, indicating its potential to complement existing methods and become a standard pre-training process for LLMs.

\end{abstract}
\begin{figure}[h]
  \centering
  \includegraphics[width=0.95\linewidth]{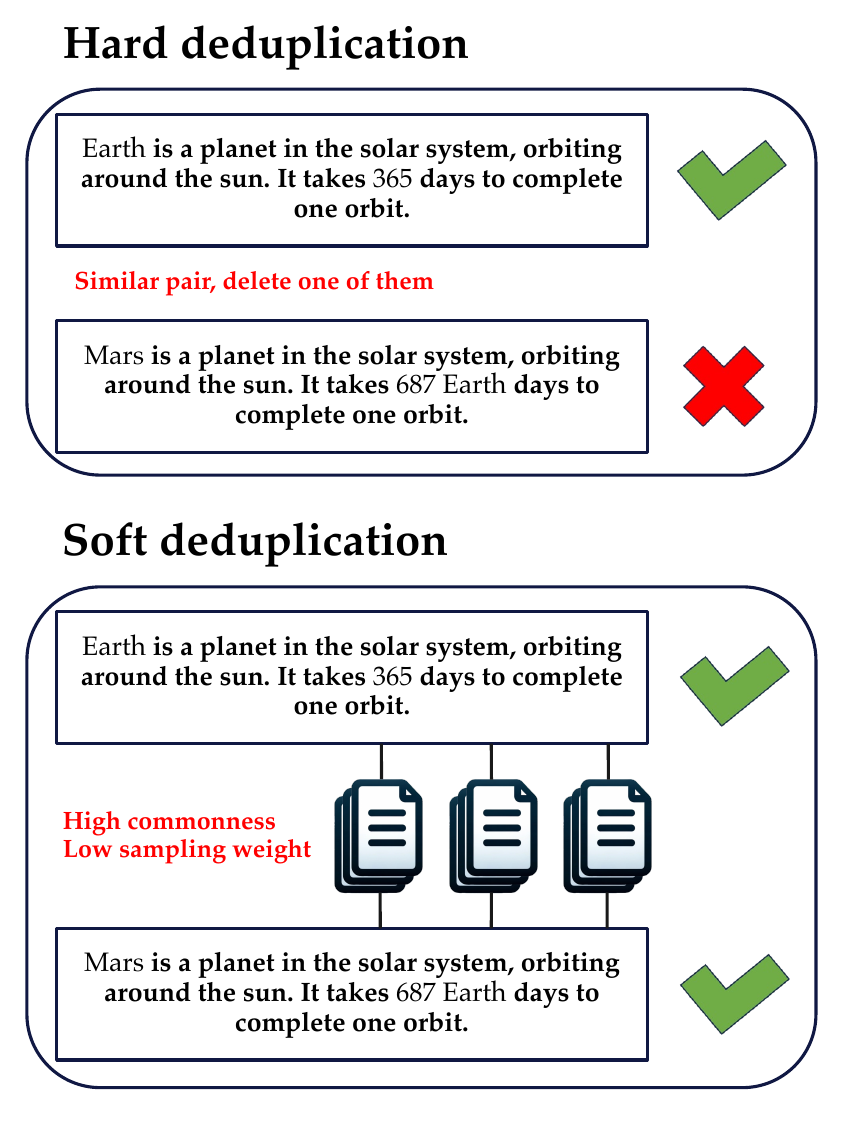}
  \caption{Hard deduplication versus soft deduplication. Hard deduplication identifies and removes duplicate samples. Soft deduplication identifies samples with high commonness, decreasing their sampling weight during training. Here, a sample refers to a document within the original corpus.}
  \label{fig:deduplication}
\end{figure}
\begin{figure*}[h]
  \centering
  \includegraphics[width=0.9\linewidth]{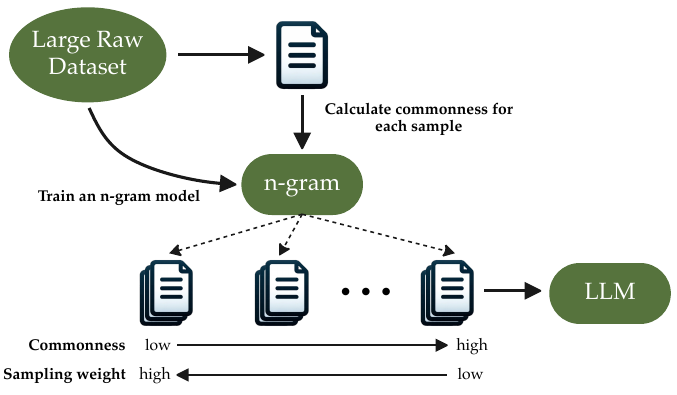}
  \caption{We aim to obtain a more balanced training set from a large raw dataset through data reweighting. Initially, we train an n-gram model using the raw dataset to calculate the commonness of each sample within the corpus. Following this, we partition the dataset and assign weights according to data commonness. Samples with higher commonness are assigned lower sampling weights, while those with lower commonness receive higher sampling weights. The weighted data is then used for the pre-training of a language model.}
  \label{fig:main_method}
\end{figure*}
\section{Introduction}

In recent years, the expansion of pre-training datasets has played a pivotal role in advancing LLMs \cite{raffel2023exploring,gao2020pile,penedo2023refinedweb}. However, a large fraction of these datasets is derived from uncurated snapshots of the internet, resulting in a significant amount of duplication. Such redundancy, particularly beyond certain levels, can severely impair the performance of LLMs \cite{hernandez2022scaling}. While repetition under specific conditions may be beneficial, the marginal gains from additional computation diminish to zero over time \cite{muennighoff2023scaling}. Thus, it is imperative to ensure that data repetition is a deliberate choice rather than an unintentional consequence. In light of this, data deduplication has emerged as a critical procedure in the management of pre-training datasets.

%Most existing data deduplication techniques can be considered as hard deduplication methods, involving the identification of redundant data and their subsequent deletion. Exact deduplication often employs suffix arrays\citep{doi:10.1137/0222058} for substring matching. However, this method struggles with subtle textual differences. For fuzzy deduplication, minhash\citep{666900} and simhash\citep{charikar2002similarity} methods are widely utilized. Yet, these approaches can mistakenly identify distinct texts as similar due to hash collisions and approximation errors. Furthermore, they fail to provide a comprehensive perspective on the overall dataset. Recent methods\citep{abbas2023semdedup,sorscher2023neural,tirumala2023d4} use pretrained embeddings for clustering to identify and discard semantically redundant data, yet they remain sensitive to hyperparameter selection and still pose a risk of information loss through direct deletion.
%Most current data deduplication strategies can be classified as hard deduplication methods. These approaches entail the identification and subsequent deletion of redundant data. Exact deduplication often employs suffix arrays \citep{doi:10.1137/0222058} for substring matching. For fuzzy deduplication, MinHash \citep{666900} and SimHash \citep{charikar2002similarity} methods are widely utilized. Recent methods \citep{abbas2023semdedup,sorscher2023neural,tirumala2023d4} use pretrained embeddings for clustering to identify and discard semantically redundant data. 
Most current data deduplication strategies can be classified as hard deduplication methods, focusing on identifying and removing redundant data. For example, MinHashLSH \citep{leskovec2020mining}, a widely utilized method \citep{cerebras2023slimpajama,penedo2023refinedweb}, approximates Jaccard similarity among samples by generating MinHash \citep{666900} signatures and using locality sensitive hashing to map these signatures into multiple buckets. Samples are considered duplicates if their MinHash values exactly match in at least one bucket, indicating they exceed a predefined similarity threshold. In the subsequent removal stage, a common practice involves clustering samples across all buckets (for instance, if samples A and B match in one bucket, and B and C in another, then A, B, and C are considered a cluster)  \citep{penedo2023refinedweb}. Within each cluster, only one sample is preserved.

%These methods face two main constraints. Firstly, the concept of the duplicates within a set of samples is symmetrical. Keeping one sample while removing the others leads to data bias, as there is no clear basis for this choice. Secondly, the degree of duplication is continuous, making it difficult to set a clear threshold for what counts as a duplicate. This can lead to a simplistic division into 'duplicated' or 'non-duplicated,' potentially oversimplifying the complexity of the data.
These methods face several principal limitations. First, the concept of duplicates within a set of samples is symmetric. Randomly retaining one sample while discarding the others may introduce bias by ignoring the differences among them. Second, setting a specific threshold for duplication presents a challenge since the degree of duplication is continuous. A high threshold might overlook near-duplicates that bear significant similarities, whereas a low threshold could result in the exclusion of valuable data. Moreover, data categorized as non-duplicates according to these thresholds are uniformly treated, despite the variations in the degree of duplication among them.

To address these limitations, we introduce a soft deduplication method (Figure \ref{fig:deduplication}). This method diverges from traditional practices by preserving the entirety of the dataset and avoids the need for setting thresholds to determine duplicates. We introduce the concept of "data commonness", a metric that quantifies the degree of duplication by measuring the occurrence probabilities of samples using an n-gram model.
Samples with high commonness are assigned a lower sampling weight, while those with low commonness receive a higher weight. This method reduces the risk of inadvertently discarding valuable data and leverages the spectrum of data duplication, offering a refined and comprehensive perspective on data deduplication.

%This work makes the following contributions:
% \begin{itemize}
%     \item We propose a novel soft data deduplication approach that enables language models to achieve baseline performance with at least 26\% fewer training steps, ultimately leading to improved performance in downstream tasks. 
%     \item %By incorporating the concept of data commonness into our deduplication process, we avoid extensive local similarity comparisons and instead adopt a global perspective. This shift not only streamlines the deduplication process but also results in a more balanced training dataset.
%     %By incorporating the concept of data commonness, we avoid extensive local similarity comparisons, prevent the deletion of valuable information, and utilize the continuity of duplication degree from a global perspective. Compared to hard deduplication methods, this approach has advantages in performance and efficiency.
%     By retaining all data and introducing the concept of data commonness, we overcome the limitations of traditional methods. Compared to hard deduplication methods, this approach has advantages in performance and efficiency.
%     \item Our method can significantly improve performance on datasets that have been rigorously deduplicated using current data management workflows. This suggests it can serve as a powerful complement to existing methods and become one of the standardized processes for pre-training LLMs.
% \end{itemize}
Our empirical analysis reveals that the proposed method enables language models to achieve baseline performance with at least 26\% fewer training steps, ultimately leading to improved performance on downstream tasks. It exhibits superior temporal efficiency and outperforms existing methods in terms of effectiveness. Significantly, even when applied to rigorously deduplicated datasets, our method still delivers substantial improvements. These results suggest that our approach can complement existing methods and can be adopted as a standard procedure in the pre-training of LLMs.
\section{Related Work}
\subsection{Data deduplication}
Research has revealed that many existing pre-training datasets contain a substantial number of duplicate samples \citep{10.1145/3133908,bandy2021addressing,penedo2023refinedweb}. To explore the impact of duplicate data on model performance, numerous studies have been conducted on both general and domain-specific datasets \citep{allamanis2019adverse,lee2022deduplicating,biderman2023pythia,xue2023repeat}. The results indicate that repetition at certain frequencies can significantly harm model performance \citep{hernandez2022scaling}. Although appropriate repetition under specific circumstances can be beneficial \citep{muennighoff2023scaling}, this should result from careful selection rather than being an unintended consequence of data duplication.

Therefore, data deduplication is crucial for pretraining large language models. Exact deduplication is typically achieved through suffix arrays \citep{doi:10.1137/0222058}. MinHash \citep{666900} and SimHash \citep{charikar2002similarity} are widely used fuzzy deduplication methods. %, which have been applied in the training of models like the LLaMA \citep{touvron2023llama} and the Falcon \citep{almazrouei2023falcon} series. 
In recent research, some studies have shifted towards semantic-based deduplication. \citet{abbas2023semdedup} and \citet{sorscher2023neural} utilize pre-trained embeddings for clustering to remove semantically redundant data. \citet{tirumala2023d4} combines both methods. 
\subsection{Data reweighting}
%Quality-based data selection is a common approach due to its significant impact on model performance \citep{longpre2023pretrainers}. GPT-3 \citep{brown2020language} utilizes original WebText as a proxy of high-quality documents and a classifier is trained to identify high-quality data. Similar methods have also been employed in PaLM \citep{chowdhery2022palm}. The data selection methods based on diversity \citep{tirumala2023d4} and importance \citep{katharopoulos2019samples} has also been proven to be valuable. Furthermore, in order to improve the model's adaptability to the target domain, it is typically necessary to select domain-specific datasets for continuous pretraining \citep{xie2023data,chen2021evaluating,lewkowycz2022solving}. However, this approach often involves manual data curation. DSIR \citep{xie2023data} introduces an automated method for selecting data that closely aligns with the target domain.
%Additionally, in the CCNet pipeline\citep{wenzek2019ccnet}, a fastText\citep{bojanowski2017enriching} linear classifier and an n-gram model are utilized to obtain high-quality data. 
Adjusting the significance of training samples through data reweighting has proven to be an effective strategy for enhancing model performance, either through modifying loss function weights or changing the sampling probabilities. Focal Loss, as introduced by \citet{lin2018focal}, employs a soft weighting scheme to allocate higher weights to more challenging samples. \citet{ren2019learning} assign weights to training samples based on the direction of their gradients. In DSIR \citep{xie2023data}, sampling based on importance weights is utilized, allowing the training data to align with the distribution of high-quality corpora such as Wikipedia. DoReMi \citep{xie2023doremi} explores an automated scheme for determining the sampling weights of different data sources.
%\subsection{Data proportion}
%Pretrained language models are typically trained on datasets with complex compositions \citep{chowdhery2022palm,brown2020language,touvron2023llama}. The partitioning and mixing ratios of the data can have a significant impact on the performance of the models \citep{hoffmann2022empirical,longpre2023pretrainers}. PaLM \citep{chowdhery2022palm} and GlaM \citep{du2022glam} determine the proportion based on downstream task performance. DoReMi \citep{xie2023doremi} can utilize group distributionally robust optimization to train a small proxy model in the domain to produce mixed proportions without prior knowledge of downstream tasks.
\section{Method}
\subsection{Hard deduplication}
Hard deduplication methods identify and remove duplicate samples. This process can be seen as partitioning the dataset $\mathcal{D}$ into numerous distinct subsets $\mathcal{D}_i$, such that $\mathcal{D}=\bigcup_{i=1}^k \mathcal{D}_i$. Each of these subsets contains samples deemed to be duplicates based on a specific similarity threshold. Within each subset $\mathcal{D}_i$, only one sample, denoted as $x_i$, is retained, while the rest are discarded. If a subset consists of only one sample, it indicates that this sample has no duplicates within the dataset.
%based on a similarity threshold. For each $\mathcal{D}_i$, a single sample is retained, denoted as $x_i$, and all other samples in $\mathcal{D}_i$ are removed.

In the context of pre-training LLMs, the fundamental training goal is to maximize the log likelihood of the training data. Incorporating hard deduplication into this process can be formulated as:
\begin{equation}
\begin{split}
%\mathcal{L} = \sum_{x\in\mathcal{D}}\sum_{i=1}^{k} I[x=x_i]\log P(x|\Theta)
\mathcal{L} &= \sum_{x\in\mathcal{D}}I(x)\log P(x|\Theta),\\
I(x)&=
\begin{cases}
1, &x\in\{x_1,x_2,\cdots,x_k\} \\
0, &\text{otherwise} \\
\end{cases}
\end{split}
\end{equation}
where $\mathcal{L}$ denotes the log likelihood function, $\Theta$ represents the model parameters. 
%$I$ is an indicator function that is active when $x$ matches the retained sample $x_i$. 
Despite its utility, hard deduplication may inadvertently omit valuable data and fail to adequately consider the degree of redundancy.
\subsection{Soft deduplication}
To address the limitations of hard deduplication, we propose a soft deduplication method. This method employs sampling weights $W(x)$, allowing for a nuanced handling of data redundancy by adjusting the influence of each sample on the model based on its commonness:
\begin{equation}
\mathcal{L} = \sum_{x \in \mathcal{D}} W(x) \cdot \log P(x | \Theta), \:W(x) \in (0,1).
\end{equation}
We assume that the sampling weight of sample $x$ can be represented as follows:
\begin{equation}
W(x)\propto\frac{1}{p(x)}.
\end{equation}
%where $p$ represents the current data distribution. The data commonness of the sample $x$ is measured by its occurrence probability $p(x)$ within the current data distribution, reflecting its degree of duplication. 
Here, $p(x)$ denotes the occurrence probability of sample $x$, serving as a direct measure of its commonness. This probability-based measure effectively captures the degree of duplication of each sample. This approach ensures that samples with higher commonness are assigned lower weights, thus mitigating the impact of duplicates without discarding potentially valuable information.
%Similar to DSIR\citep{xie2023data}, we assume that the sampling weight of the sample $x$ can be represented as follows:
% \begin{equation}
% W(x)=C\cdot\frac{p(x)}{q(x)}
% \end{equation}
% where $p$ represents the desired data distribution, $q$ represents the current data distribution and $C$ is a constant. The data commonness of the sample $x$ is measured by its occurrence probability $q(x)$ within the current data distribution. Our desired data distribution is characterized by maximal diversity, which necessitates adhering to the principle of maximum entropy. This leads us to conclude that $p$ ought to be a uniform distribution. Consequently, this premise simplifies the equation to: 
% \begin{equation}
% W(x)=C'\cdot\frac{1}{q(x)}.
% \end{equation}
\subsection{Implementation of commonness calculation}
%In practical implementation, we utilize an n-gram model to efficiently calculate the commonness of each data sample (Figure \ref{fig:main_method}). This process consists of 3 steps.
In practical implementation, we leverage an n-gram model to process data, achieving high temporal efficiency in calculating the commonness of each data sample (Figure \ref{fig:main_method}). This process consists of three steps.
\begin{enumerate}
  \item \textbf{Tokenization.} The n-gram model assumes that the appearance of a word is determined by the previous $n-1$ words. The first step is to tokenize the original corpus. We use the same tokenizer as the pre-training models for consistency.
  \item \textbf{Train n-gram model.} %In the training process of an n-gram model ($n=4$), maximum likelihood estimation is used to calculate the probability of each n-gram. "$n=4$" is the empirical choice made after our early experiments. To alleviate the issue of data sparsity, we employ the Kneser-Ney smoothing technique\citep{NEY19941}. The KenLM toolkit\footnote{https://kheafield.com/code/kenlm} is utilized to accomplish this step.
  In the training process of an n-gram model (where $n=4$), maximum likelihood estimation is used to calculate the probability of each n-gram. We empirically choose $n=4$ after conducting early experiments. To alleviate the issue of data sparsity, we employ the Kneser-Ney smoothing technique \citep{NEY19941}. We utilize the KenLM toolkit\footnote{https://kheafield.com/code/kenlm} to accomplish this step.
  \item \textbf{Calculate commonness.} We utilize the obtained n-gram model to compute the commonness (measured by the occurrence probability) for each data sample. For a given $x$ containing $N$ tokens, 
  \begin{equation}
  p(x) = \left(\prod_{i=1}^{N} P(w_i |w_{i-1},\ldots,w_{i-n+1})\right)^{\frac{1}{N}}.
  \end{equation}
  By employing the geometric mean, the influence of sample length can be eliminated. 
  %\begin{equation}
  % \begin{split}
  % &PPL(x) = q(x)^{-\frac{1}{N}} \\
  % &=e^{-\frac{1}{N}\sum\limits_{i=1}^{N}\log P(w_i |w_{i-1},\ldots,w_{i-n+1})}
  % \end{split}
  % \end{equation}
  %where $N$ represents the number of tokens.
\end{enumerate}
\subsection{Approximate sampling for large-scale data}
%We can obtain the sampling weight of a given $x$:
% \begin{equation}
% q(x)=e^{\frac{T}{N}\sum\limits_{i=1}^{N}\log P(w_i |w_{i-1},\ldots,w_{i-n+1})}\label{eq:ppl}
% \end{equation}
% \begin{equation}
% \begin{split}
% W(x)&=C'\cdot q(x)^{-\frac{T}{N}} \\
% &=C'\cdot e^{-\frac{T}{N}\sum\limits_{i=1}^{N}\log P(w_i |w_{i-1},\ldots,w_{i-n+1})}\label{eq:ppl}.
% \end{split}
% \end{equation}
% A hyperparameter $T$ is introduced to adjust the disparity between the maximum and minimum sampling weights.

Due to the vast volume of data, directly assigning individual sampling weights to each data point is impractical. To overcome this, we introduce an approximate method for data sampling that segments $M$ samples into $K$ categories. This process initiates by sorting the $M$ samples in ascending order of data commonness, followed by dividing the dataset into $K$ distinct segments according to $K$ quantiles. For the $k$-th segment, the sampling weight $W_{k}$ is determined by the $k$-th quantile, $p_k$, as follows:
\begin{equation}
W_{k} = C \cdot \left(\frac{1}{p_k}\right)^T\label{eq:ppl}
%C = \frac{1}{\sum_{j=1}^{K} \left(\frac{1}{p(q_j)}\right)^T}
\end{equation}
where $T$ is a hyperparameter that adjusts the sampling weight and $C$ is a normalization constant, which ensures that the sum of the weights across all segments equals one. 
% \begin{equation}
% W_{Q_k}= W(x), \text{ where }x=Quartile_k.
% \end{equation}
% \begin{equation}
% W(x)=p(x)^{-T}\label{eq:ppl}.
% \end{equation}
\section{Experimental Setup}
%This section aims to provide a detailed description of our experimental setup. We employ a decoder-only model architecture and conduct experiments on three corpora of different qualities. Moreover, we explore various data partitioning numbers and weight allocation strategies. The evaluation includes the perplexity of the models on the test datasets and their performance on downstream tasks.
\subsection{Datasets}
We conduct experiments on different versions of the Common Crawl dataset, which is a comprehensive and publicly accessible collection of data obtained through web crawling. %For each dataset, we randomly sample 40 billion tokens for training.%The first one, from RedPajama \citep{together2023redpajama}, involves the original Common Crawl data undergoing processing through the CCNet pipeline \citep{wenzek2019ccnet}. This preprocessing includes a paragraph-level deduplication process and the deployment of a linear classifier aimed at identifying and selecting texts of superior quality. Additionally, we incorporate the SlimPajama Common Crawl dataset \citep{cerebras2023slimpajama} into our analysis. The SlimPajama dataset represents a further refined iteration of the RedPajama corpus, boasting enhanced data cleansing procedures and the implementation of MinHashLSH \citep{leskovec2020mining} for more effective deduplication, thus resulting in a more streamlined dataset. The third dataset is the Falcon RefinedWeb \citep{penedo2023refinedweb}, which similarly adopts rigorous deduplication and filtering methodologies. For each dataset, we randomly sample 40 billion tokens for training.

\noindent\textbf{RedPajama CommonCrawl} is a subset of the RedPajama dataset \citep{together2023redpajama}. It involves the original Common Crawl data undergoing processing through the CCNet pipeline \citep{wenzek2019ccnet}. This dataset has been subjected to paragraph-level deduplication; however, it has not undergone rigorous deduplication procedures.
%This preprocessing includes a paragraph-level deduplication process and the deployment of a linear classifier aimed at identifying and selecting texts of superior quality. 

\noindent\textbf{SlimPajama CommonCrawl} is a subset of the SlimPajama dataset \citep{cerebras2023slimpajama}. The SlimPajama dataset represents a further refined iteration of the RedPajama corpus, boasting enhanced data cleansing procedures and the implementation of MinHashLSH \citep{leskovec2020mining} for more effective deduplication.

\noindent\textbf{Falcon RefinedWeb} is introduced as a pre-training dataset for the Falcon series \citep{penedo2023refinedweb,almazrouei2023falcon}. It undergoes rigorous deduplication processes using exact matching and MinHashLSH.
\subsection{Model training}
 In the experiments, we employ the same model architecture as the LLaMA \citep{touvron2023llama} series. Our models are configured with 1.3B parameters, incorporating 16 attention heads and 24 layers. The hidden size is set to 2048, and the dimension of feed-forward network is 5504. Previous research has demonstrated the feasibility of pre-training validation on models of this scale \citep{tirumala2023d4,xie2023doremi}. All models are trained from scratch to 40B tokens. The batch size is 512, and the training sequence length is 1024. The learning rate is decayed from 2e-4 to 2e-5. 
\subsection{Baselines}
Our primary baseline is defined by directly training on a dataset that has been randomly sampled to encompass 40B tokens. In our study, we implement the SoftDedup method across all three datasets, facilitating a comparative analysis between our proposed technique and the established baseline for each dataset. Furthermore, for experiments conducted on the RedPajama CommonCrawl dataset, the SlimPajama CommonCrawl, which employs MinHashLSH for deduplication directly on it, is considered a hard deduplication baseline.
%We propose two main baselines for our study. The first baseline involves training directly on the randomly sampled dataset of 40 billion tokens. Additionally, we consider clustering as a commonly used method for data partitioning\citep{abbas2023semdedup,sorscher2023neural,tirumala2023d4}. To obtain vector representations of the documents, we train a language model with 120M parameters on the same dataset. We then utilize the k-means algorithm for data clustering. Previous research suggests that the number of clusters should roughly correspond to the square root of the total dataset size \citep{abbas2023semdedup}. To improve the uniformity of the training data, we assign equal weights to the data after clustering and train models on this adjusted dataset.
\subsection{Evaluation metrics}
We evaluate the models by measuring their perplexity on the test sets and their few-shot performance on downstream tasks.

\noindent\textbf{Test set perplexity.} Our test sets come from the Pile \citep{gao2020pile} and SlimPajama \citep{cerebras2023slimpajama}. The Pile test set consists of 22 subsets, including BookCorpus, DM Mathematics, and others. SlimPajama also includes 7 subsets, such as Common Crawl, C4, and GitHub. We measure the perplexity of the models on each sample and report the average for each subset. We investigate data leakage and remove the contaminated samples. Specifically, if a sample in the training set has more than 50 tokens of overlap with a sample in the test set, the former will be removed from the training set.

\noindent\textbf{Downstream task accuracy.} In order to further evaluate the performance of the models, we measure their accuracy on 12 downstream tasks. 
%We report the average few-shot accuracy on 12 downstream tasks. 
The tasks cover the models' abilities in reading comprehension (SQuADv2 \citep{rajpurkar2018know}, Trivia QA \citep{JoshiTriviaQA2017}), commonsense reasoning (ARC easy and challenge \citep{Clark2018ThinkYH}, WinoGrande \citep{sakaguchi2019winogrande}, HellaSwag \citep{zellers2019hellaswag}, PIQA \citep{Bisk2020}, Social IQa \citep{sap2019social}), world knowledge (WebQuestions \citep{berant-etal-2013-semantic}, NQ Open \citep{lee-etal-2019-latent}), and contextual understanding (LAMBADA standard and openai \citep{paperno2016lambada}). The evaluation of downstream tasks is primarily accomplished through the utilization of the lm-evaluation-harness \citep{eval-harness}.
\subsection{Hyperparameter impact analysis}
%We adjust the number of data partitions and modify the weight allocation method to explore the effects of these different strategies.
In exploring the impact of hyperparameters on our method, we focus on two key hyperparameters: the number of data partitions ($K$) and the weight parameter ($T$).

Our experiments involves varying levels of data partition granularity by dividing the dataset into 10, 20, 50, and 100 segments. Regarding weight assignment, we modify the hyperparameter $T$ within Equation \ref{eq:ppl} to alter weight disparities. We investigate three configurations that result in maximum-minimum weight differences of approximately 2-fold, 5-fold, and 10-fold, respectively. A larger disparity exerts a greater suppression on data with higher commonness.
%Additionally, we conduct experiments using the DoReMi method to automatically determine the optimal weights, and the weights obtained from our approach are used as initial weights.
\begin{figure*}[h]
  \centering
  \subfloat[The Pile test set]{\includegraphics[width=0.33\textwidth]{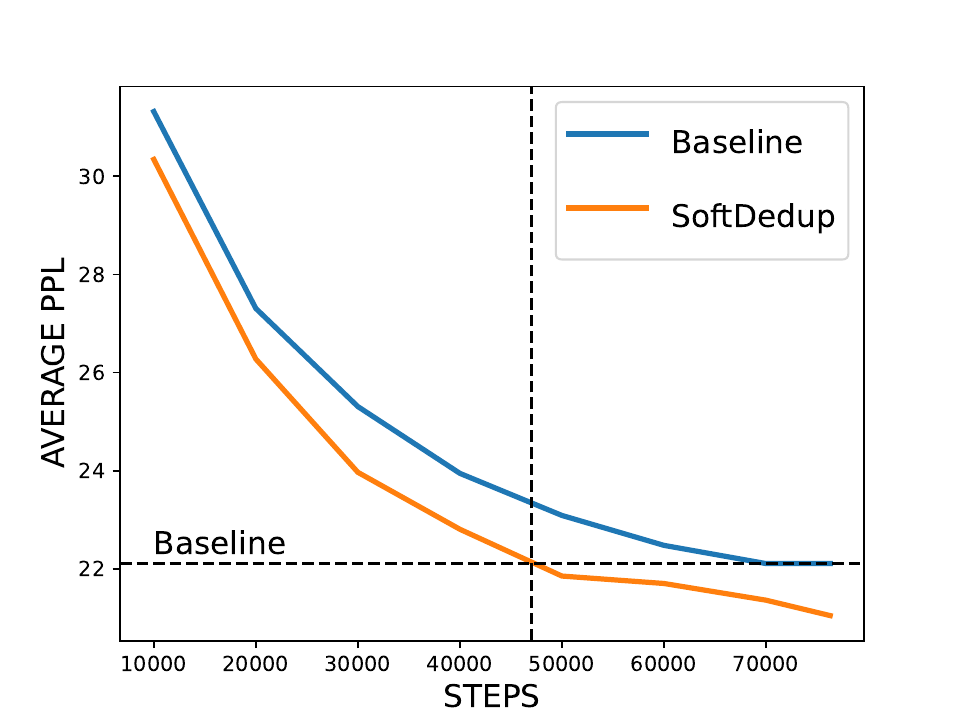}\label{fig:redcc_pile}}
  %\hfill
  \subfloat[SlimPajama test set]{\includegraphics[width=0.33\textwidth]{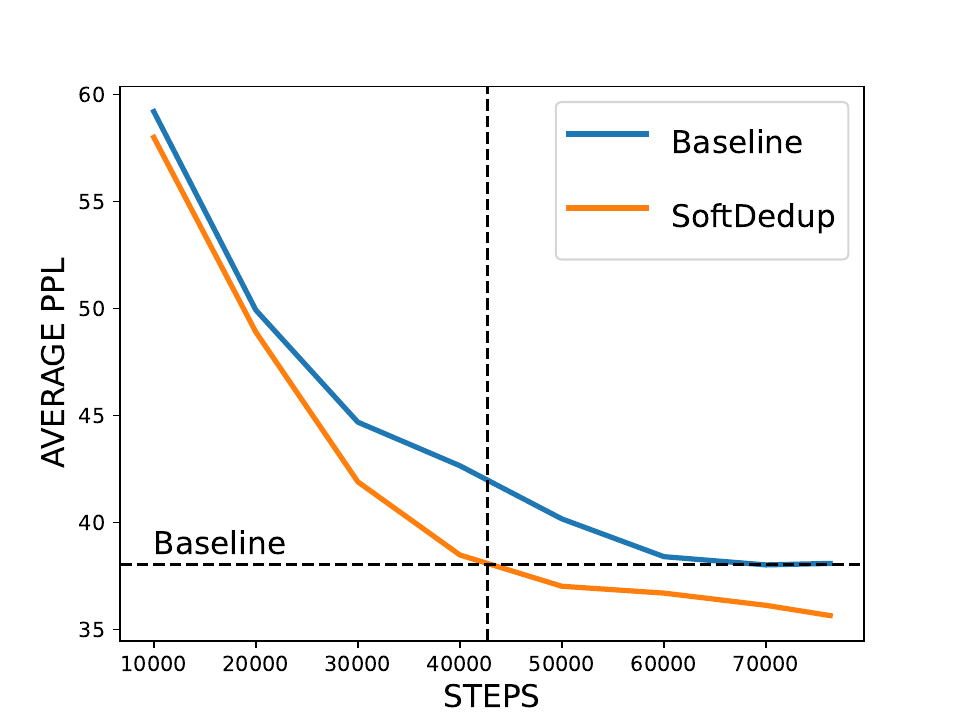}\label{fig:redcc_slimpajama}}
  %\hfill
  \subfloat[downstream tasks]{\includegraphics[width=0.33\textwidth]{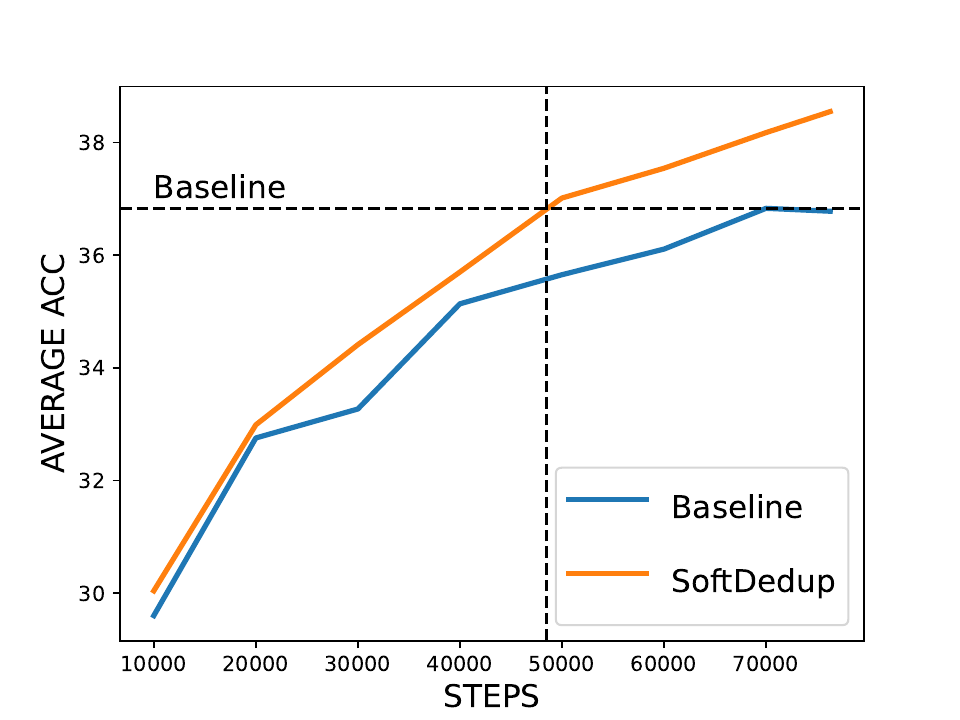}\label{fig:redcc_downstream}}
  \caption{Performance evaluation results of models trained on the RedPajama CommonCrawl dataset. Figures \ref{fig:redcc_pile} and \ref{fig:redcc_slimpajama} display the average perplexity on the Pile and SlimPajama test sets, respectively. Figure \ref{fig:redcc_downstream} illustrates the average accuracy on various downstream tasks. Our methodology involves a data partitioning number of 20 and a 10-fold weight disparity between the maximum and minimum weights. Baseline refers to direct training.}
  \label{fig:red}
\end{figure*}
\begin{table*}
\centering
\begin{tabular}{l|c|cc|cc}
\hline
%&\multicolumn{2}{c|}{ \textbf{RedPajama CC}} &\multicolumn{2}{c|}{ \textbf{SlimPajama CC}} &\multicolumn{2}{c}{ \textbf{Falcon RW}} \\
%\hline
 \textbf{Task}&\textbf{Baseline}&\textbf{HardDedup}&\textbf{Difference}&\textbf{SoftDedup}&\textbf{Difference}\\
\hline
 NQ Open (1-shot) &  4.13 &  4.6  &  +0.47&  \textbf{5.37} & \textbf{+1.24}  \\
 SQuADv2 (1-shot) &  11.51   & 12.95  & +1.44 &  \textbf{14.66} &  \textbf{+3.15}  \\
  Trivia QA (1-shot) &  15.89   & \textbf{17.71}  &  \textbf{+1.82} &  16.39 &  +0.5  \\
 WebQuestions (1-shot) &  3.3   &  \textbf{5.71} &  \textbf{+2.41}  &  3.4 & +0.1  \\
 LAMBADA openai (1-shot) &  46.07   &  43.64 & -2.43   &  \textbf{48.52} & \textbf{+2.45}  \\
 LAMBADA standard (1-shot) &  36.91   &  37.65 & +0.74   &  \textbf{40.89} & \textbf{+3.98}  \\
 PIQA (1-shot) &  65.34   &  66 &  +0.66 &  \textbf{66.7} &  \textbf{+1.36}  \\
 Social IQa (1-shot) &  88   & 87.9  &  -0.1 &  \textbf{89.6} &  \textbf{+1.6} \\
 WinoGrande (1-shot) &  52.88   &  53.99 &  +1.11 &  \textbf{54.38} & \textbf{+1.5} \\
 HellaSwag (1-shot) &  34.93   & 35.54  & +0.61  &  \textbf{36.51} & \textbf{+1.58} \\
 ARC easy (2-shot) &  57.24   & 57.79  & +0.55 &  \textbf{59.89} &  \textbf{+2.65} \\
 ARC challenge (2-shot) &  25.17   &  25.09 & -0.08  &  \textbf{26.28} &  \textbf{+1.11} \\
 Average & 36.78  &  37.38 & +0.6  & \textbf{38.55} & \textbf{+1.77} \\
\hline
\end{tabular}
\caption{\label{tab:main}Performance comparison of models on downstream tasks using soft and hard deduplication methods.} 
\end{table*}

\begin{figure*}[h]
  \centering
  \subfloat[The Pile test set]{\includegraphics[width=0.33\textwidth]{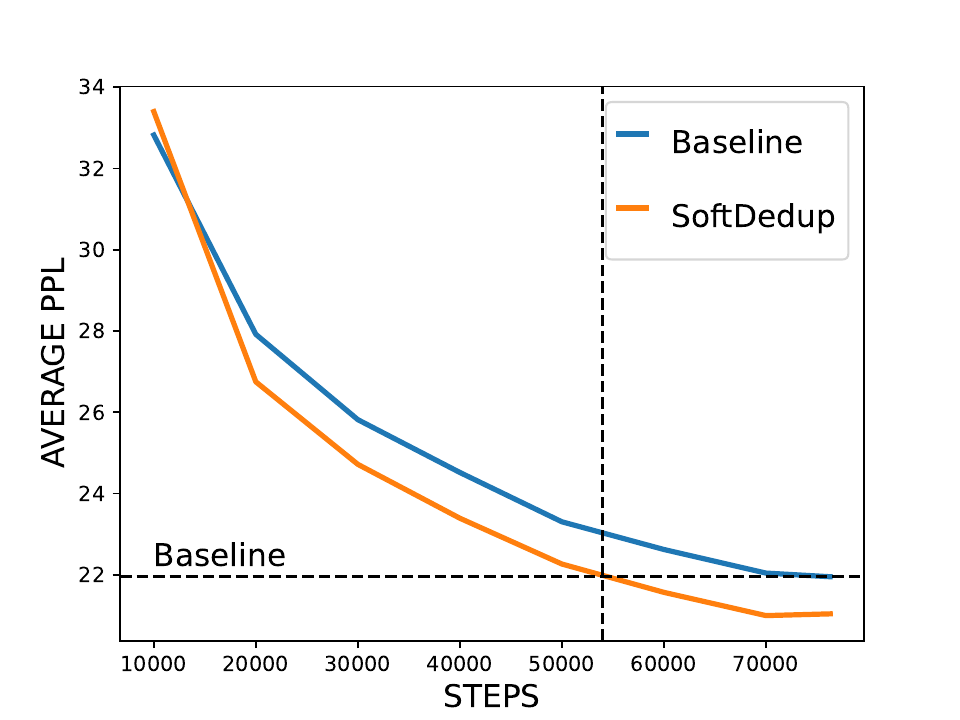}\label{fig:slimcc_pile}}
  %\hfill
  \subfloat[SlimPajama test set]{\includegraphics[width=0.33\textwidth]{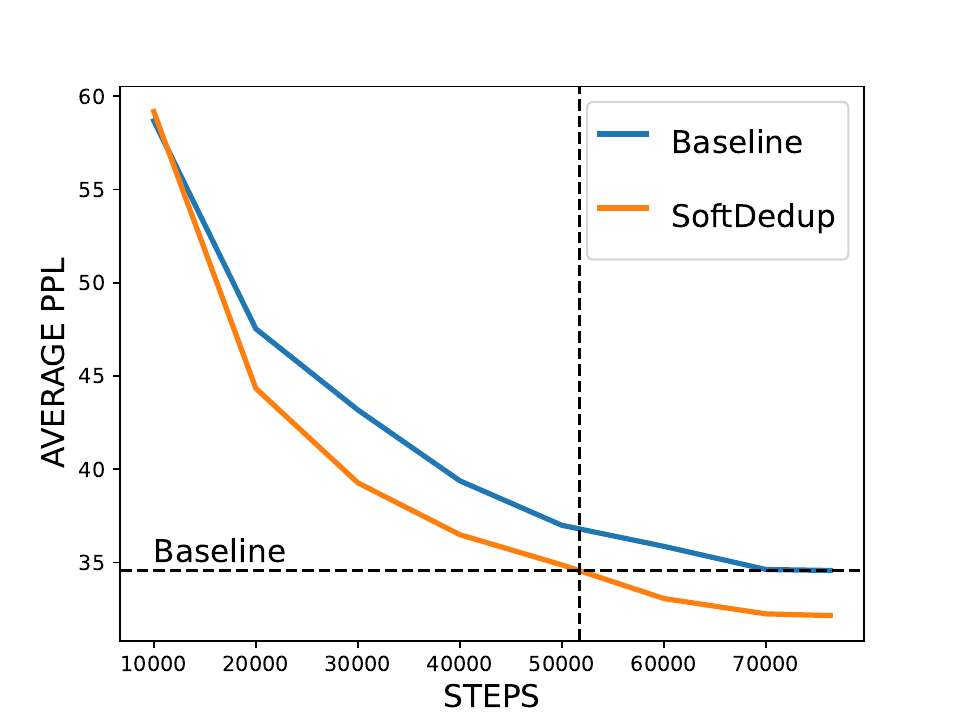}\label{fig:slimcc_slimpajama}}
  %\hfill
  \subfloat[downstream tasks]{\includegraphics[width=0.33\textwidth]{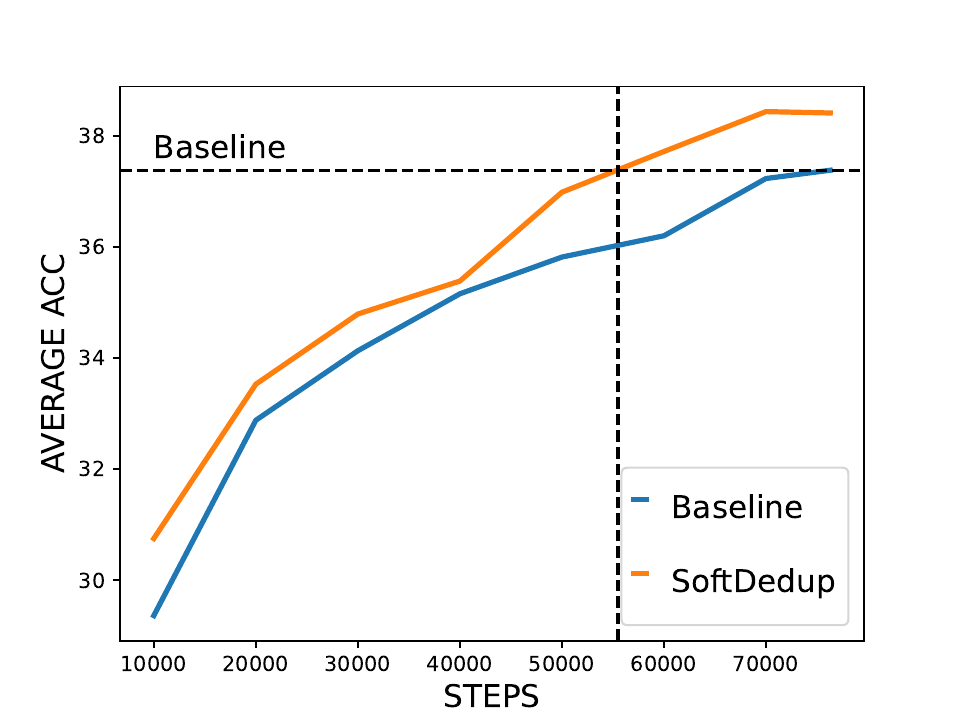}\label{fig:slimcc_downstream}}
  \caption{Performance evaluation results of models trained on the SlimPajama CommonCrawl dataset. Figures \ref{fig:slimcc_pile} and \ref{fig:slimcc_slimpajama} display the average perplexity on the Pile and SlimPajama test sets, respectively. Figure \ref{fig:slimcc_downstream} illustrates the average accuracy on various downstream tasks. Our methodology involves a data partitioning number of 20 and a 10-fold weight disparity between the maximum and minimum weights. Baseline refers to direct training.}
  \label{fig:slim}
\end{figure*}
\begin{figure*}[htbp]
  \centering
  \subfloat[The Pile test set]{\includegraphics[width=0.33\textwidth]{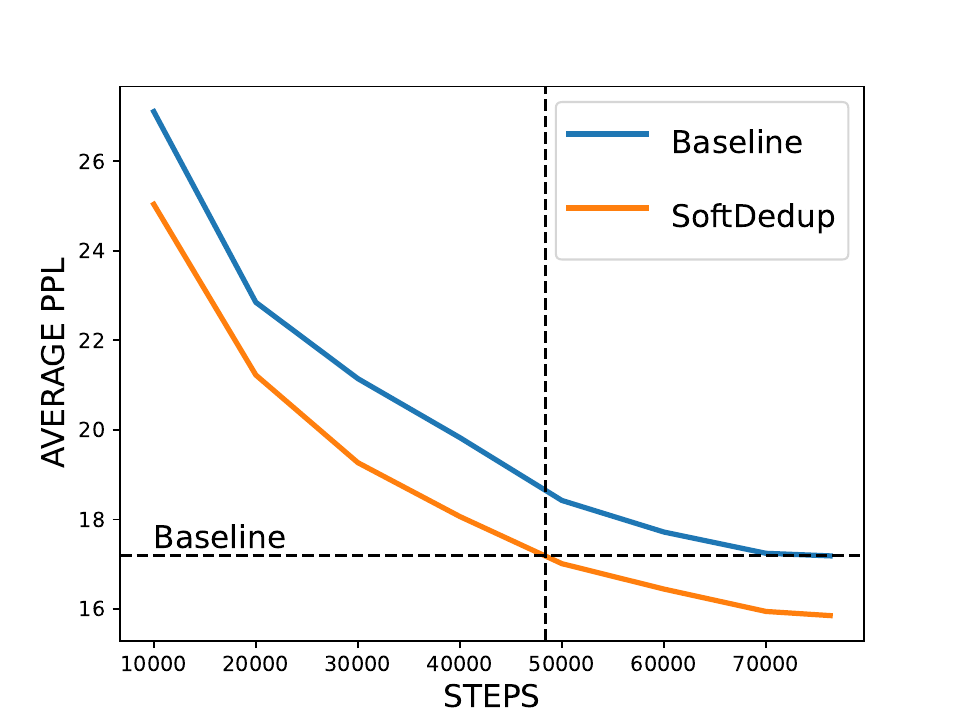}\label{fig:falcon_pile}}
  %\hfill
  \subfloat[SlimPajama test set]{\includegraphics[width=0.33\textwidth]{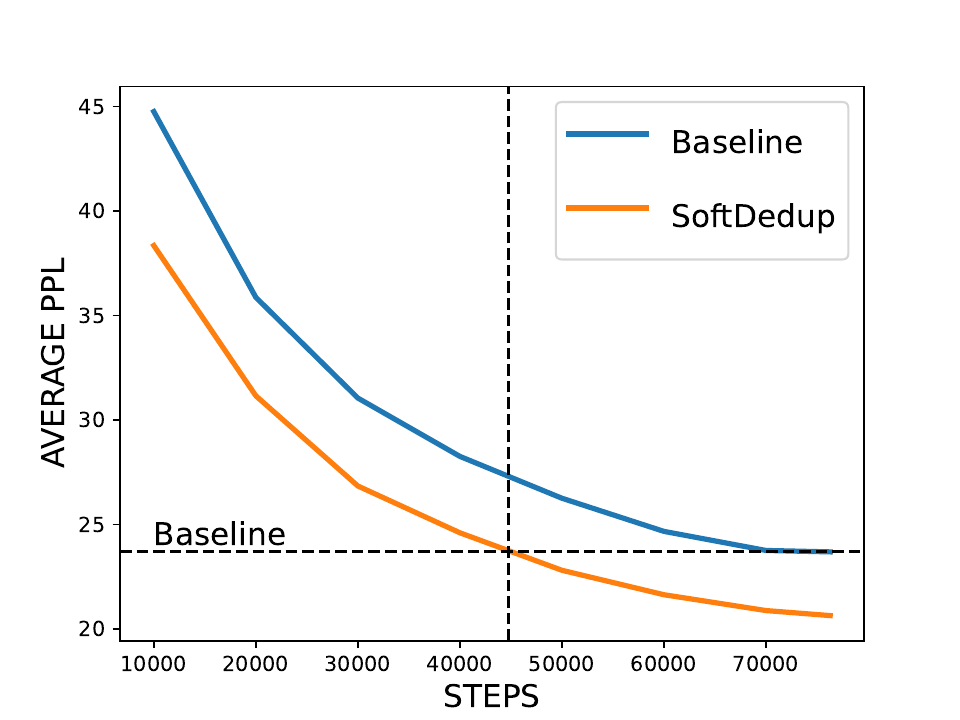}\label{fig:falcon_slimpajama}}
  %\hfill
  \subfloat[downstream tasks]{\includegraphics[width=0.33\textwidth]{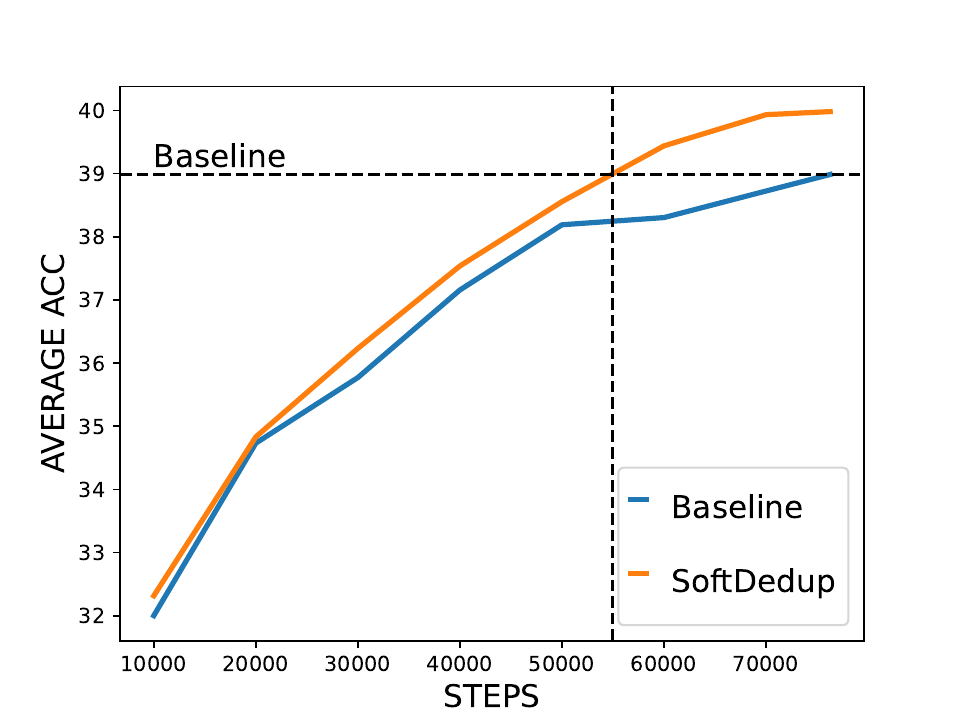}\label{fig:falcon_downstream}}
  \caption{Performance evaluation results of models trained on the Falcon RefinedWeb dataset.}
  \label{fig:falcon}
\end{figure*}

\begin{figure*}[h]
  \centering
  \subfloat[The Pile test set]{\includegraphics[width=0.33\textwidth]{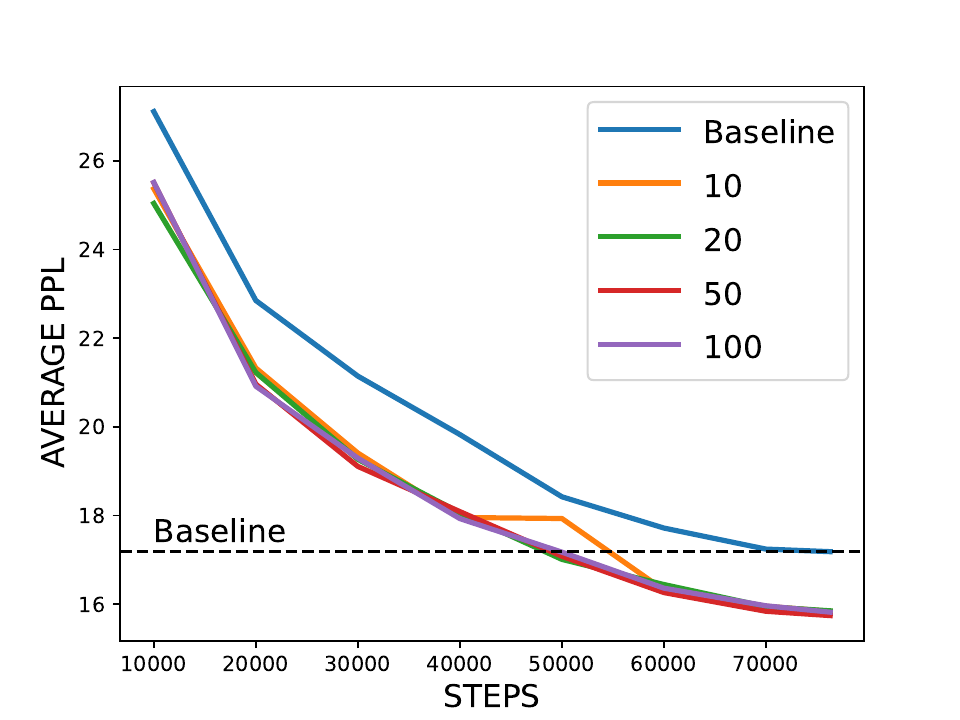}\label{fig:partition_num_pile}}
  %\hfill
  \subfloat[SlimPajama test set]{\includegraphics[width=0.33\textwidth]{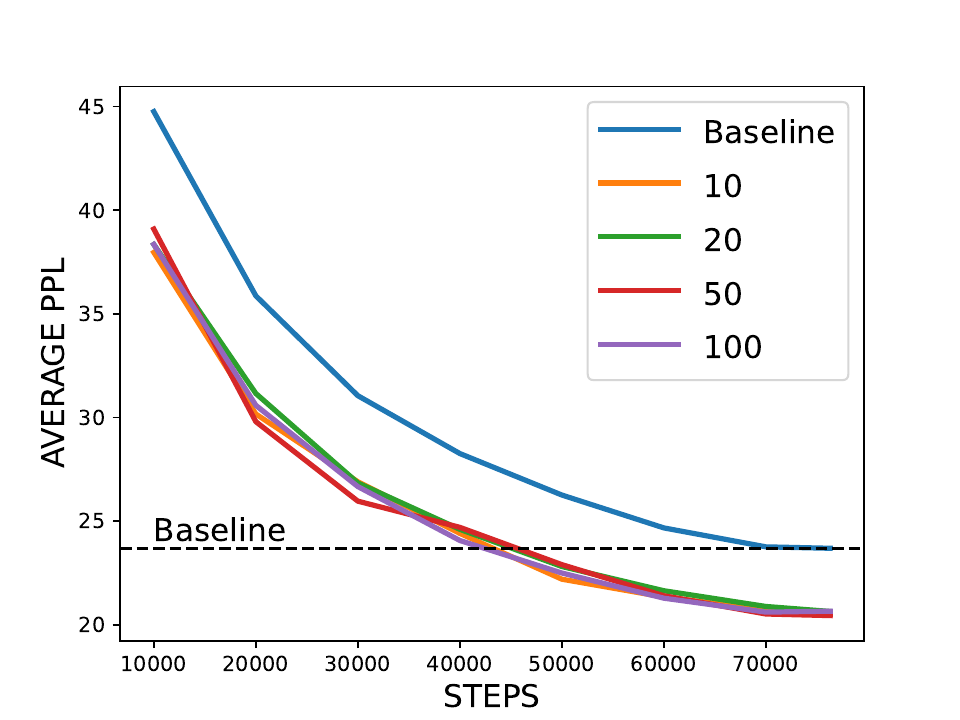}\label{fig:partition_num_slimpajama}}
  %\hfill
  \subfloat[downstream tasks]{\includegraphics[width=0.33\textwidth]{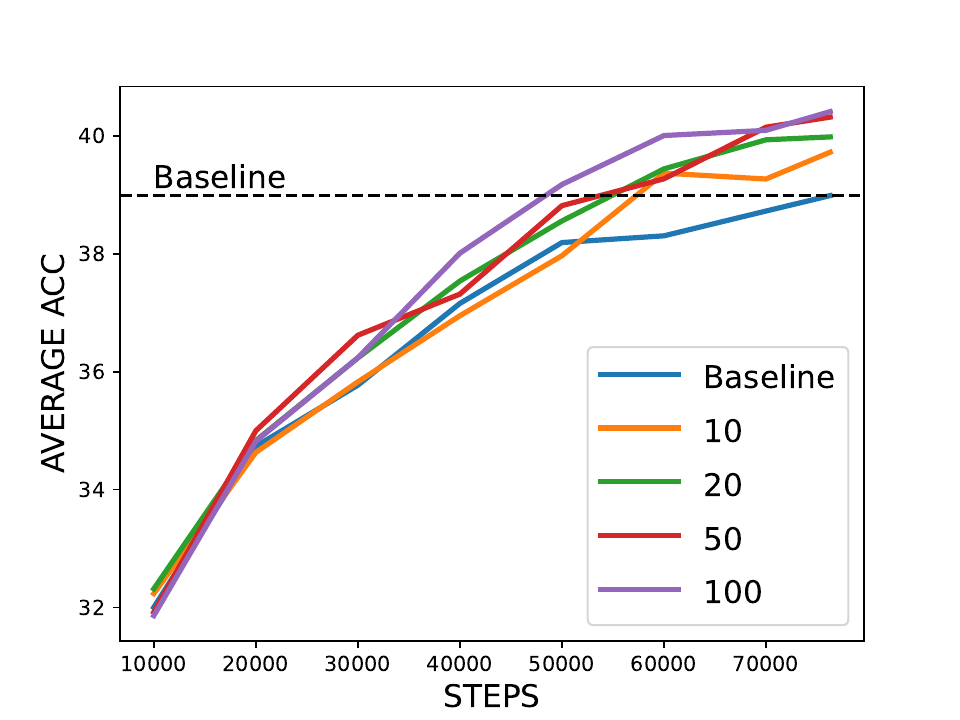}\label{fig:partition_num_downstream}}
  \caption{The effect of data partition number on model performance. The models are trained on the Falcon RefinedWeb dataset, applying a 10-fold weight disparity between maximum and minimum weights. Data partitions are set at 10, 20, 50, and 100.}
  \label{fig:partition_num}
\end{figure*}
\begin{figure*}
  \centering
  \subfloat[The Pile test set]{\includegraphics[width=0.33\textwidth]{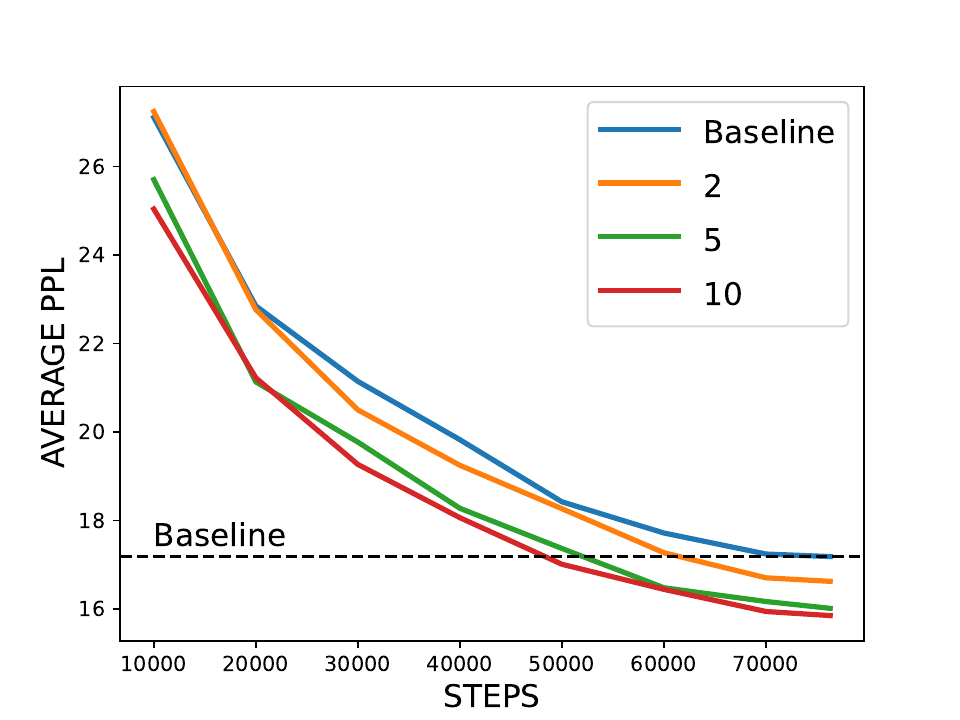}\label{fig:weight_pile}}
  \hfill
  \subfloat[SlimPajama test set]{\includegraphics[width=0.33\textwidth]{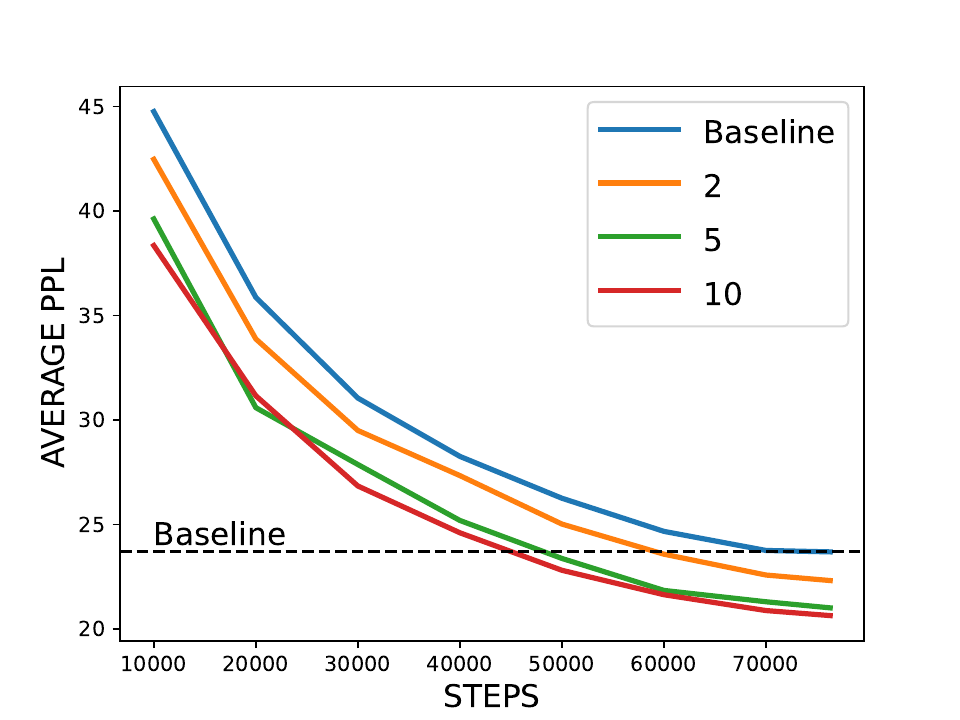}\label{fig:weight_slimpajama}}
  \hfill
  \subfloat[downstream tasks]{\includegraphics[width=0.33\textwidth]{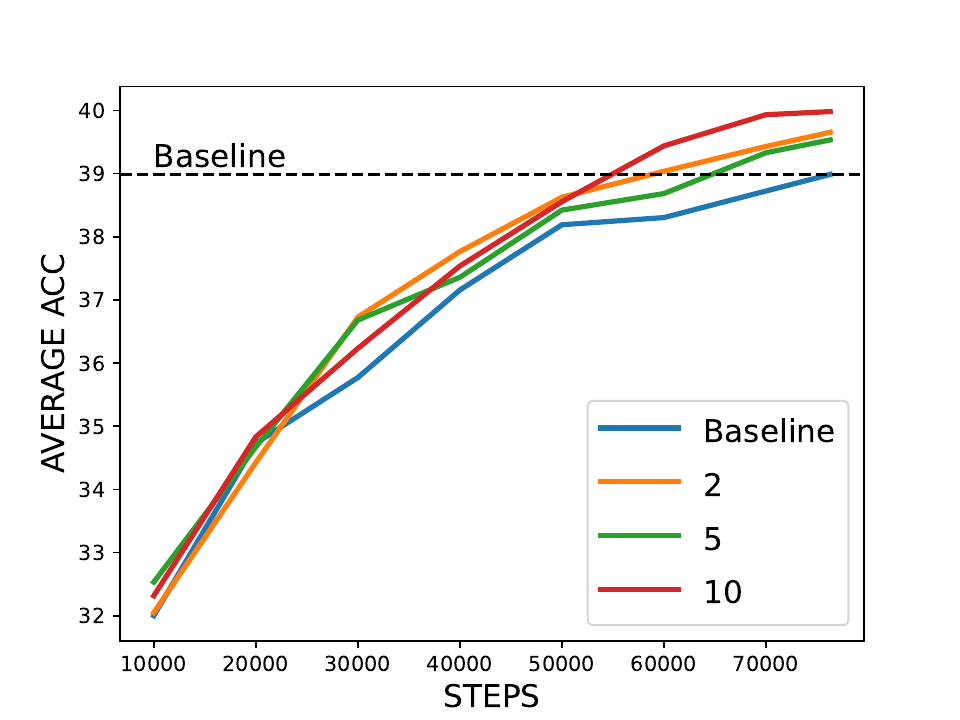}\label{fig:weight_downstream}}
  \caption{Evaluation results of models under different weighting disparities, including maximum-minimum weight differences of approximately 2-fold, 5-fold, and 10-fold. The models are trained on the Falcon RefinedWeb dataset, and our method involves a data partitioning number of 20.}
  \label{fig:weight}
\end{figure*}
\section{Results}
In this section, we provide a detailed report of the results under various experimental settings. 
\subsection{Enhanced performance and efficiency in language model pre-training}
To verify the effectiveness of our soft deduplication method, we conduct experiments on the RedPajama CommonCrawl dataset, which has not subjected to meticulous deduplication. Our findings indicate a significant improvement with our method compared to the direct training baseline, as illustrated in Figure \ref{fig:red}. 

%\textbf{Perplexity on test sets.}
Our approach consistently outperforms the baseline in terms of average perplexity across all evaluated datasets. Specifically, on the Pile test set, our method enables models to achieve baseline perplexity within 50,000 iterations, saving nearly 30,000 training steps. Furthermore, models continue to improve, ultimately reaching a lower perplexity, as shown in Figure \ref{fig:redcc_pile}. Similar advancements are observed in the SlimPajama test set, confirming our method's effectiveness (Figure \ref{fig:redcc_slimpajama}). Additionally, we report the average perplexity for each subset upon completion of training (Appendices \ref{sec:appendix_pile_ppl} and \ref{sec:appendix_slimpajma_ppl}). Our method enables the models to yield performance improvements across the majority of the test subsets.

%\textbf{Accuracy on downstream tasks.} 
In our evaluation of downstream tasks, our method outperforms the baseline in accuracy. It accelerates learning on the RedPajama dataset, achieving baseline performance nearly 30,000 steps sooner and improving average accuracy by 1.77\% at the end of training, as shown in Figure \ref{fig:redcc_downstream}. Detailed scores for each individual task at the final training checkpoint are delineated in Table \ref{tab:main}. Our approach yields improvements in all evaluated tasks.

In summary, our experiments on the RedPajama CommonCrawl dataset substantiate that the soft deduplication method is capable of reducing perplexity and enhancing the accuracy of downstream tasks more efficiently compared to the baseline model. Such accelerated convergence is crucial for pre-training large language models, considering the significant costs associated with training duration and resource utilization.
\subsection{Surpassing hard deduplication in effectiveness}
In the experiments carried out using the RedPajama CommonCrawl dataset, we also contrast the SoftDedup method against traditional hard deduplication techniques (refer to Table \ref{tab:main}). Considering that the SlimPajama dataset originates from the RedPajama dataset, refined through MinHashLSH deduplication, we employ models trained on the SlimPajama CommonCrawl dataset as the hard deduplication baseline.

The evaluation results of models on various downstream tasks reveal our method's superior performance over both the hard deduplication technique and the direct training baseline. In detail, while the hard deduplication method surpasses the direct training baseline in nine out of twelve tasks, showing an average increase in accuracy of 0.6\%, our SoftDedup method demonstrates more consistent and significant improvements. It outperforms the direct training baseline across all evaluated tasks, achieving an average accuracy enhancement of 1.77\%. These findings underscore the advantages over conventional deduplication methods in enhancing downstream task performance.
%To facilitate a more effective comparison with existing methodologies, we conduct a comparative analysis of models trained on two datasets: the RedPajama CommonCrawl dataset and and its refined counterpart, the SlimPajama CommonCrawl dataset (refer to Table \ref{tab:main}). 

%The SlimPajama CommonCrawl dataset has undergone refinement via the MinHashLSH deduplication process, adhering to a Jaccard similarity threshold of 0.8. The results show a clear improvement in model performance after deduplication. Specifically, the baseline perplexity on the Pile test set decreases from 22.11 to 21.96, and on the SlimPajama test set, it is reduced from 38.09 to 34.58. Concurrently, there is an increase in average downstream task accuracy from 36.78\% to 37.38\%.

%The results obtained from deploying our soft deduplication methodology on the RedPajama CommonCrawl dataset substantiate its efficacy, outperforming the existing deduplication techniques. Notably, the method achieves an enhanced downstream task accuracy rate of 38.55\%, which is a marked improvement over the 37.38\% accuracy rate obtained by the baseline model trained on the SlimPajama CommonCrawl dataset. This enhancement is also corroborated by the model's performance on the Pile test set. Our methodology mitigates the impact of data redundancy through judicious adjustments in data distribution and exhibits the potential to supplant current methodologies.

\subsection{A powerful complement to existing techniques}
To further assess the effectiveness of our method when applied in sequence with extant hard deduplication processes, we conduct experiments on the SlimPajama CommonCrawl and Falcon RefinedWeb datasets, which have undergone stringent deduplication processes (as shown in Figures \ref{fig:slim} and \ref{fig:falcon}). 

%\textbf{Perplexity on test sets.}
In the evaluations conducted on the Pile and SlimPajama test sets, our method exhibits consistent superiority over the baseline models. Notably, our models achieve equivalent baselines in perplexity with a reduction of 26\% to 39\% in the number of required training steps. Additionally, the ultimate performance of the models demonstrates a tangible enhancement, as evidenced by the results displayed in Figures \ref{fig:slimcc_pile}, \ref{fig:slimcc_slimpajama}, \ref{fig:falcon_pile}, and \ref{fig:falcon_slimpajama}.
In terms of accuracy on downstream tasks, Figures \ref{fig:slimcc_downstream} and \ref{fig:falcon_downstream} highlight the training efficiency achieved by our models. It is particularly noteworthy that our method reaches baseline performance with around 20,000 fewer training steps. We report the detailed scores in Appendices \ref{sec:appendix_pile_ppl}, \ref{sec:appendix_slimpajma_ppl}, and \ref{sec:appendix_downstream}.

In summary, even when applied to already deduplicated datasets, our method significantly enhances training efficiency and effectiveness. This underscores its capability to address the shortcomings of current deduplication techniques. Specifically, our approach reweights the data to reflect varying levels of duplication, thus avoiding one-size-fits-all solutions. This integration has the potential to become a standard practice in the pre-training of large language models.
%In summary, when applied to already deduplicated datasets, our method still significantly enhances training efficiency and effectiveness, underscoring its ability to compensate for the shortcomings of current deduplication methods. Particularly, it reweights data to reflect varying duplication levels, avoiding blanket solutions. This integration could become a standard in large language model pre-training.
\subsection{Finer data partitioning for improved downstream task performance}
In Figure \ref{fig:partition_num}, we illustrate the impact of different numbers of data partitions on model performance. We argue that investigating methods to further enhance the training effectiveness of higher-quality data is a more critical concern. Therefore, our experiments are conducted on the Falcon RefinedWeb dataset.

In evaluations conducted on both the Pile and SlimPajama test sets, models exhibit negligible variations in average perplexity across a range of data partition counts, specifically 10, 20, 50, and 100. This observation indicates that perplexity, as a metric, demonstrates relatively low sensitivity to changes in the granularity of data partitioning.

In contrast, we observe that as the granularity of data partitioning increases, the accuracy of the language model in downstream tasks also improves. As demonstrated in Figure \ref{fig:partition_num_downstream}, there is a clear correlation between the number of data partitions and the model's accuracy. This indicates that finer-grained data partitioning can make the training data more balanced, thereby enhancing performance in downstream tasks.
\subsection{Effects of sampling weight disparities on model performance}
Figure \ref{fig:weight} presents the outcomes of experimental investigations into the effects of varying disparities between maximum and minimum weights assigned to different data partitions. The methodology employed ensures a consistent ascending order in the allocation of weights, with greater disparities indicating a more pronounced suppression of data with high commonness. 

Experiments conducted utilizing disparities in the maximum to minimum weight ratios of 2-fold, 5-fold, and 10-fold reveal a consistent trend: increased disparities between the maximum and minimum weights lead to a reduction in average model perplexity. Although slight variations are observed in the performance outcomes for downstream tasks, the experiments demonstrate that the largest weight disparity consistently facilitates the most optimal model performance.
\subsection{Cost of data reweighting}
%In the detailed series of experiments described in this work, our research has conclusively demonstrated that the SoftDedup methodology consistently improves training efficiency. Specifically, for models comprising 1.3 billion parameters, the implementation of SoftDedup has enabled the attainment of equivalent baseline performance while necessitating at least a 26\% decrease in the number of required training steps. This efficiency improvement is quantitatively substantial, resulting in a reduction of approximately 930 GPU hours in our experimental setup, utilizing exclusively 32G V100 GPUs.
The computational processes of n-gram training and commonness calculation are executed solely on CPU resources. For a 40B token corpus, the n-gram training procedure (with $n=4$) requires 4 CPU cores for 5 hours, followed by computing data commonness using 4 CPU cores in 2 hours. Compared to the substantial costs of GPU conservation (at least 930 V100 GPU hours in our experiments), these expenses can be considered negligible. This underscores the efficiency of SoftDedup and the feasibility of its implementation in resource-constrained environments.
\section{Conclusion}
In this study, we introduce a soft deduplication method that effectively addresses the primary limitations associated with traditional hard deduplication methods. Unlike its predecessors, this approach retains all samples of data while reallocating sampling weights according to data commonness. Experimental analyses demonstrate that this technique can significantly expedite the training process for large language models, evidenced by a reduction of over 26\% in the number of training steps required. The proposed method surpasses existing deduplication techniques in effectiveness and can serve as a valuable complement to these methods. Due to its low operational cost and superior efficiency, we advocate for the integration of this soft deduplication approach with traditional hard deduplication methods as a standard practice in the pre-training phase of large language models.
\section*{Limitations}
Due to current limitations in computational resources, the extension of SoftDedup to larger-scale models will be deferred to future research endeavors. Moreover, future studies will seek to conduct a more comprehensive evaluation of the method's effectiveness across various mixed data sources.
% \subsection{References}

% \nocitep{Ando2005,andrew2007scalable,rasooli-tetrault-2015}

% The \LaTeX{} and Bib\TeX{} style files provided roughly follow the American Psychological Association format.
% If your own bib file is named \texttt{custom.bib}, then placing the following before any appendices in your \LaTeX{} file will generate the references section for you:
% \begin{quote}
% \begin{verbatim}
% \bibliography{custom}
% \end{verbatim}
% \end{quote}

% You can obtain the complete ACL Anthology as a Bib\TeX{} file from %\url{https://aclweb.org/anthology/anthology.bib.gz}.
% To include both the Anthology and your own .bib file, use the following instead of the above.
% \begin{quote}
% \begin{verbatim}
% \end{verbatim}
% \end{quote}

% Please see Section~\ref{sec:bibtex} for information on preparing Bib\TeX{} files.

% \subsection{Appendices}

% Use \verb|\appendix| before any appendix section to switch the section numbering over to letters. See Appendix~\ref{sec:appendix} for an example.
%\section*{Acknowledgements}
%\clearpage
\bibliography{anthology,custom}
\appendix
\section{Appendix}
\subsection{Average perplexity for each subset in the Pile test set}
\label{sec:appendix_pile_ppl}
\begin{table*}
\centering
\begin{tabular}{l|cc|cc|cc}
\hline
&\multicolumn{2}{c|}{ \textbf{RedPajama CC}} &\multicolumn{2}{c|}{ \textbf{SlimPajama CC}} &\multicolumn{2}{c}{ \textbf{Falcon RW}} \\
\hline
 \textbf{Subset}&\textbf{Baseline}&\textbf{SoftDedup}&\textbf{Baseline}&\textbf{SoftDedup}&\textbf{Baseline}&\textbf{SoftDedup}\\
\hline
 Pile-CC          &  17.79 &  \textbf{17.20} &  20.61 &  \textbf{19.74} &  13.66 &  \textbf{13.49} \\
 YoutubeSubtitles &  23.59 &  \textbf{22.54} &   \textbf{23.40} &  25.83 &  18.56 &  \textbf{17.45} \\
 PhilPapers       &  15.74 &  \textbf{14.59} &  15.27 &  \textbf{14.51} &  15.03 &  \textbf{14.05} \\
 HackerNews       &  31.32 &  \textbf{29.81} &  29.68 &  \textbf{28.84} &  21.13 &  \textbf{20.17} \\
 Enron Emails     &  48.23 &  \textbf{46.50} &  47.16 &  \textbf{41.81} &  32.53 &  \textbf{31.26} \\
 EuroParl         &  60.96 &  \textbf{55.16} &  60.26 &  \textbf{55.72} &  51.06 &  \textbf{40.48} \\
 Ubuntu IRC       &  20.53 &  \textbf{19.48} &  27.16 &  \textbf{26.20} &  75.61 &  \textbf{71.47} \\
 BookCorpus2      &  16.22 &  \textbf{16.14} &  15.27 &  \textbf{14.46} &  11.67 &  \textbf{11.44} \\
 NIH ExPorter     &  10.92 &  \textbf{10.78} &  \textbf{10.65} &  10.72 &  \textbf{10.46} &  10.64 \\
 OpenSubtitles    &  14.45 &  \textbf{13.78} &  14.05 &  \textbf{13.67} &  13.83 &  \textbf{13.60} \\
 Gutenberg(PG-19) &  \textbf{19.67} &  19.77 &  18.71 &  \textbf{17.75} &  18.31 &  \textbf{16.58} \\
 DM Mathematics   &  6.51 &  \textbf{6.44} &  \textbf{6.47} &  6.60 &  6.01 &  \textbf{5.86} \\
 Wikipedia        &  13.94 &  \textbf{13.71} &  12.85 &  \textbf{12.65} &  10.70 &  \textbf{10.45} \\
 OpenWebText2     &  21.97 &  \textbf{21.10} &  27.16 &  \textbf{25.44} &  17.00 &  \textbf{16.19} \\
 Github           &  56.03 &  \textbf{52.95} &  55.98 &  \textbf{53.65} &  32.36 &  \textbf{26.30} \\
 FreeLaw          &  \textbf{9.86} &  10.13 &  10.37 &  \textbf{10.26} &  13.36 &  \textbf{12.93} \\
 USPTO Backgrounds&  9.59 &  \textbf{9.29} &  9.60 &  \textbf{9.42} &  9.19 &  \textbf{8.97} \\
 Books3           &  \textbf{15.69} &  16.02 &  14.82 &  \textbf{14.37} &  11.06 &  \textbf{10.65} \\
 PubMed Abstracts &  \textbf{8.75} &  8.79 &  \textbf{8.49} &  8.67 &  \textbf{8.85} &  8.99 \\
 StackExchange    &  31.76 &  \textbf{29.44} &  29.83 &  \textbf{27.44} &  17.84 &  \textbf{15.62} \\
 ArXiv            &  17.82 &  \textbf{16.99} &  \textbf{17.85} &  18.14 &  16.17 &  \textbf{14.94} \\
 PubMed Central   &  13.44 &  \textbf{12.36} &  12.60 &  \textbf{12.19} &  \textbf{12.06} &  12.77 \\
\hline
\end{tabular}
\caption{\label{tab:ppl_pile}Average perplexity for each subset in the Pile test set.}
\end{table*}

%  arxiv &  17.82 &  16.99 &  17.85 &  18.14 &  16.17 &  14.94 \\
%  bc2 &  16.22 &  16.14 &  15.27 &  14.46 &  11.67 &  11.44 \\
%  b3 &  15.69 &  16.02 &  14.82 &  14.37 &  11.06 &  10.65 \\
%  dmm &  6.51 &  6.44 &  6.47 &  6.60 &  6.01 &  5.86 \\
%  ee &  48.23 &  46.50 &  47.16 &  41.81 &  32.53 &  31.26 \\
%  ep &  60.96 &  55.16 &  60.26 &  55.72 &  51.06 &  40.48 \\
%  fl &  9.86 &  10.13 &  10.37 &  10.26 &  13.36 &  12.93 \\
%  github &  56.03 &  52.95 &  55.98 &  53.65 &  32.36 &  26.30 \\
%  gp19 &  19.67 &  19.77 &  18.71 &  17.75 &  18.31 &  16.58 \\
%  hn &  31.32 &  29.81 &  29.68 &  28.84 &  21.13 &  20.17 \\
%  ne &  10.92 &  10.78 &  10.65 &  10.72 &  10.46 &  10.64 \\
%  os &  14.45 &  13.78 &  14.05 &  13.67 &  13.83 &  13.60 \\
%  owt2 &  21.97 &  21.10 &  27.16 &  25.44 &  17.00 &  16.19 \\
%  pp &  15.74 &  14.59 &  15.27 &  14.51 &  15.03 &  14.05 \\
%  pcc &  17.79 &  17.20 &  20.61 &  19.74 &  13.66 &  13.49 \\
%  pa &  8.75 &  8.79 &  8.49 &  8.674020879 &  8.85 &  8.99 \\
%  pc &  13.44 &  12.36 &  12.60 &  12.19 &  12.06 &  12.77 \\
%  s &  31.76 &  29.44 &  29.83 &  27.44 &  17.84 &  15.62 \\
%  uirc &  20.53 &  19.48 &  27.16 &  26.20 &  75.61 &  71.47 \\
%  ub &  9.59 &  9.29 &  9.60 &  9.42 &  9.19 &  8.97 \\
%  wiki &  13.94 &  13.71 &  12.85 &  12.65 &  10.70 &  10.45 \\
%  ys &  23.59 &  22.54 &  23.40 &  25.83 &  18.56 &  17.45 \\
In Table \ref{tab:ppl_pile}, we provide a detailed report on the average perplexity for each subset within the Pile test set. For models trained on the RedPajama CommonCrawl dataset, our method results in improvements across 18 out of 22 subsets. For models trained on the SlimPajama CommonCrawl dataset, our method leads to improvements in 17 subsets. For models trained on the Falcon RefinedWeb, improvements are observed in 19 subsets. Due to the exceedingly small number of documents in the Ubuntu IRC subset, we exclude it from the calculation of the average perplexity on the Pile test set.
\subsection{Average perplexity for each subset in the SlimPajama test set}
\label{sec:appendix_slimpajma_ppl}
\begin{table*}
\centering
\begin{tabular}{l|cc|cc|cc}
\hline
&\multicolumn{2}{c|}{ \textbf{RedPajama CC}} &\multicolumn{2}{c|}{ \textbf{SlimPajama CC}} &\multicolumn{2}{c}{ \textbf{Falcon RW}} \\
\hline
 \textbf{Subset}&\textbf{Baseline}&\textbf{SoftDedup}&\textbf{Baseline}&\textbf{SoftDedup}&\textbf{Baseline}&\textbf{SoftDedup}\\
\hline
 Commoncrawl  &  9.43 &  \textbf{9.28} &  9.19 &  \textbf{9.16} &  10.23 &  \textbf{10.20} \\
 C4           &  16.84 &  \textbf{16.25} &  16.46 &  \textbf{16.06} &  13.95 &  \textbf{13.81} \\
 GitHub       &  90.00 &  \textbf{82.28} &  81.11 &  \textbf{77.59} &  28.02 &  \textbf{20.98} \\
 Books        &  \textbf{16.03} &  16.25 &  14.62 &  \textbf{13.91} &  11.89 &  \textbf{11.34} \\
 ArXiv        &  15.86 &  \textbf{15.29} &  15.21 &  \textbf{14.57} &  14.75 &  \textbf{13.43} \\
 Wikipedia    &  88.40 &  \textbf{82.05} &  77.44 &  \textbf{67.77} &  68.83 &  \textbf{58.69} \\
 StackExchange&  30.10 &  \textbf{28.15} &  28.00 &  \textbf{26.01} &  18.14 &  \textbf{16.01} \\
\hline
\end{tabular}
\caption{\label{tab:ppl_slimpajama}Average perplexity for each subset in the SlimPajama test set.} 
\end{table*}

%  arxiv &  15.86018423 &  15.29495353 &  15.21329435 &  14.57239444 &  14.75243262 &  13.42838592 \\
%  book &  16.02919359 &  16.25843995 &  14.61747335 &  13.91961239 &  11.88769006 &  11.34073036 \\
%  c4 &  16.83795743 &  16.25299281 &  16.4615613 &  16.06868871 &  13.95030617 &  13.81082819 \\
%  cc &  9.428575453 &  9.280957942 &  9.192487296 &  9.165244573 &  10.23303043 &  10.20273772 \\
%  github &  89.99749539 &  82.28382844 &  81.10606038 &  77.5956705 &  28.02202073 &  20.98330811 \\
%  s &  30.09938552 &  28.15557447 &  28.00251647 &  26.00813642 &  18.14112831 &  16.00610423 \\
%  wiki &  88.3984406 &  82.05743141 &  77.4373389 &  67.76605199 &  68.8270391 &  58.69434935 \\
In Table \ref{tab:ppl_slimpajama}, we provide a detailed report on the average perplexity for each subset within the SlimPajama test set. Our method has led to improvements across nearly all subsets.
\subsection{Accuracy for each downstream task}
\label{sec:appendix_downstream}
\begin{table*}
\centering
\begin{tabular}{l|cc|cc|cc}
\hline
&\multicolumn{2}{c|}{ \textbf{RedPajama CC}} &\multicolumn{2}{c|}{ \textbf{SlimPajama CC}} &\multicolumn{2}{c}{ \textbf{Falcon RW}} \\
\hline
 \textbf{Task}&\textbf{Baseline}&\textbf{SoftDedup}&\textbf{Baseline}&\textbf{SoftDedup}&\textbf{Baseline}&\textbf{SoftDedup}\\
\hline
 NQ Open (1-shot) &  4.13 &  \textbf{5.37} &  4.6 &  \textbf{4.79} &  \textbf{4.18} &  3.85 \\
 SQuADv2 (1-shot) &  11.51 &  \textbf{14.66} &  12.95 &  \textbf{17.25} &  26.39 &  \textbf{29.95} \\
  Trivia QA (1-shot) &  15.89 &  \textbf{16.39} &  \textbf{17.71} &  17.64 &  12.24 &  \textbf{13.3} \\
 WebQuestions (1-shot) &  3.3 &  \textbf{3.4} &  \textbf{5.71} &  3.59 &  1.08 &  \textbf{3.05} \\
 LAMBADA openai (1-shot) &  46.07 &  \textbf{48.52} &  43.64 &  \textbf{44.65} &  44.3 &  \textbf{46.57} \\
 LAMBADA standard (1-shot) &  36.91 &  \textbf{40.89} &  37.65 &  \textbf{38.83} &  39.72 &  \textbf{40.79} \\
 PIQA (1-shot) &  65.34 &  \textbf{66.7} &  66 &  \textbf{66.76} &  72.31 &  \textbf{72.47} \\
 Social IQa (1-shot) &  88 &  \textbf{89.6} &  87.9 &  \textbf{88.4} &  89.2 &  \textbf{89.3} \\
 WinoGrande (1-shot) &  52.88 &  \textbf{54.38} &  53.99 &  \textbf{55.25} &  54.7 &  54.7 \\
 HellaSwag (1-shot) &  34.93 &  \textbf{36.51} &  35.54 &  \textbf{36.8} &  \textbf{40.43} &  40.35 \\
 ARC easy (2-shot) &  57.24 &  \textbf{59.89} &  57.79 &  \textbf{60.44} &  58.12 &  \textbf{59.43} \\
 ARC challenge (2-shot) &  25.17 &  \textbf{26.28} &  25.09 &  \textbf{26.54} &  25.17 &  \textbf{26.02} \\
\hline
\end{tabular}
\caption{\label{tab:downstream}Accuracy for each downstream task.} 
\end{table*}
In Table \ref{tab:downstream}, we provide a detailed report on the accuracy for each downstream task. For models trained on the RedPajama CommonCrawl dataset, our method has led to improvements across all tasks. For models trained on the SlimPajama CommonCrawl and Falcon RefinedWeb datasets, our approach has also resulted in accuracy improvements on the majority of tasks.
\end{document}